\documentclass{esannV2}
\usepackage{amssymb,amsmath,array}
\usepackage[a4paper, total={122mm, 193 mm},  left=44mm, top=52mm]{geometry}
\usepackage[T1]{fontenc}
\usepackage{hyperref}
\usepackage{url}
\usepackage{booktabs}
\usepackage{amsfonts}
\usepackage{nicefrac}
\usepackage{microtype}
\usepackage[table]{xcolor}
\usepackage{import}         
\usepackage{graphicx}       
\usepackage{wrapfig}        
\usepackage{lipsum}
\usepackage{amsmath}
\usepackage{bm}             
\usepackage{algorithm}
\usepackage{multirow}
\usepackage{algpseudocode}

\algnewcommand{\Initialize}[1]{%
  \State \textbf{Initialize:} \hspace*{\algorithmicindent}\parbox[t]{.8\linewidth}{\raggedright #1}
}
\algrenewcommand\algorithmicrequire{\textbf{Input:}}

\begin{document}
\setlength{\abovedisplayskip}{1pt}
\setlength{\belowdisplayskip}{1pt}
\title{Explore Reinforced: Equilibrium Approximation with Reinforcement Learning}

\author{Ryan Yu$^1$, Mateusz Nowak$^2$, Qintong Xie$^2$, Michelle Yilin Feng$^1$ and Peter Chin$^2$
\vspace{.3cm}\\
1 - Boston University - Computer Science Department \\ 
Boston, MA 02215 - United States
\vspace{.1cm}\\
2 - Dartmouth College - Thayer School of Engineering \\
Hanover, NH 03755 - United States}

\maketitle
\begin{abstract}
  Current approximate Coarse Correlated Equilibria (CCE) algorithms struggle with equilibrium approximation for games in large stochastic environments but are theoretically guaranteed to converge to a strong solution concept. In contrast, modern Reinforcement Learning (RL) algorithms provide faster training yet yield weaker solutions. We introduce Exp3-IXrl - a blend of RL and game-theoretic approach, separating the RL agent’s action selection from the equilibrium computation while preserving the integrity of the learning process. We demonstrate that our algorithm expands the application of equilibrium approximation algorithms to new environments. Specifically, we show the improved performance in a complex and adversarial cybersecurity network environment - the Cyber Operations Research Gym - and in the classical multi-armed bandit settings.
\end{abstract}
\section{Introduction}
Reinforcement Learning (RL) is a goal-oriented machine learning paradigm that primarily focuses on how agents act in an environment to maximize cumulative reward. In game theory, a Nash Equilibrium describes a strategy in which no player can gain by unilaterally changing their strategy if the strategies of the others remain unchanged \cite{Myerson1991}. In most real-world scenarios, which consist of dynamic complex environments in multi-step situations, the equilibrium (or equilibria) is computationally intractable \cite{daskalakis2005three}.

Game theory and RL both focus on decision-making within uncertain environments, with RL learning an optimal policy that maximizes some notion of a reward and game theory analyzing strategic interactions among players and predicting equilibrium outcomes of these interactions \cite{GT1991}. Combining these two disciplines has led to the development of agents that not only adapt to multifaceted, changing, or uncertain settings but also anticipate and strategically react to the actions of other agents, thereby enhancing the effectiveness of decision-making systems in various applications \cite{rl-gt-survey}. 

In this paper, we describe a new technique that adapts the powerful, computational feasibility of the Exponential-weight algorithm for Exploration and Exploitation (EXP3) algorithm to approximate the coarse correlated equilibrium (CCE) \cite{exp3}. We combine EXP3’s high probability successor, EXP3-IX \cite{exp3ix}, with heuristics of the Local Best Response (LBR) algorithm \cite{Lis2016EqilibriumAQ} for sequential games and the power of reinforcement learning, to create this game-theoretic guide for reinforcement learning models.
\section{Background}

A simultaneous-move task can be represented as a stochastic game. We define a basic, fully observable, N-player stochastic game as $(\mathcal{S}, \mathcal{H}, \{\mathcal{A}_i\}_{i \in N}, \mathcal{T}, \{\mathcal{R}_i\}_{i \in N}, \gamma)$, where $\mathcal{S}$ is set of all states shared by all $N$ players, $\mathcal{H}$ is the maximum number of time steps, $\mathcal{A}_i$ is the action space for player $i$ and $\mathcal{A}:= \mathcal{A}_1 \times \dots \times \mathcal{A}_N$ is a set of all valid actions, $\mathcal{T}: (\mathcal{S} \times \mathcal{A}) \rightarrow \mathcal{S}$ is a transition function, $\mathcal{R}_i:(\mathcal{S} \times A \times \mathcal{S}) \rightarrow \mathbb{R}$ is a reward function for player $i$ and $\gamma \in \left[0,1\right]$ is a discount factor.

The notion of an equilibrium provides a strong learning objective in multi-agent settings. The most popular equilibrium is the Nash Equilibrium (NE). While approximating NE is ideal, it was shown to be PPAD-complete even in 2p0s games \cite{PPAD}. Weaker, more computationally feasible, equilibrium forms can be approximated using no-regret learning. In this study, we approximate a coarse correlated equilibrium (CCE) \cite{exp3}. A CCE, $\sigma$, is defined as: $$\forall \; i, s_i' \quad \mathbb{E}_{s \sim \sigma} \ c_i(s) \leq \mathbb{E}_{s \sim \sigma} \ c_i(s_i', s_{-i})$$ where $i$ represents a player, $s_i'$ represents a strategy different from the recommended strategy, $s$, and $c_i$ represents the cost of following a strategy. Recently, the lower bound on number of iterations for the convergence of $\epsilon$-CCE for a three-player extensive-form game was proven to be $2^{\log_2^{1-o(1)}(|G|)}$, where $|G|$ is the size of the game \cite{pmlr-v247-peng24a}, while in two-player games with no chance moves, a social-welfare maximizing extensive-form CCE can be computed in polynomial time \cite{farina2020coarse}.

Traditionally, \textit{regret} $R_T$ at time step $T$ and a probabilistic bound, which deals with the uncertainty introduced by learners in multi-armed bandits, called \textit{pseudo-regret} $\hat{R}_T$ is defined as:
$$R_T = \sum_{t=1}^T \ell_{t, I_t} - \min_{i \in [K]} \sum_{t=1}^T \ell_{t,i} \text{ ,  and  } \hat{R}_T = \max_{i \in [K]} \mathbb{E} \left[\sum_{t=1}^T \ell_{t, I_t} - \sum_{t=1}^T \ell_{t,i}\right],$$
where the player has an action space of size $K$, $\ell_{t,k}$ represents the loss experienced at time step $t$ for action $i \in [K]$, and $I_t$ defines a forecaster's choice \cite{Cesa-Bianchi2006,exp3ix, Bubeck2012}. No-regret learning measures the difference in loss compared to the best single action in hindsight. In this study, we utilize EXP-IX, demonstrating no-regret learning with a high probability \cite{exp3ix}. 

\section{Related Work}

\subsection{Modern reinforcement learning algorithms}

Reinforcement Learning (RL) has seen significant advancements through algorithms like Deep Q-Networks (DQN) \cite{mnih2015human} and Policy Gradient methods \cite{sutton1999policy}, achieving remarkable results in Atari games and robotics. DQN introduced deep neural networks to approximate Q-values, enabling breakthroughs in complex environments. Policy gradient algorithms refine policies via gradient ascent on estimated returns but can suffer from instability and inefficiency due to high-variance gradients. TRPO \cite{schulman2015trust} and PPO \cite{schulman2017proximal} address this by stabilizing updates with relative entropy constraints.

Multi-Agent Reinforcement Learning (MARL) is closely related to game theory and repeated games. Algorithms such as multi-agent deterministic policy gradient (MADDPG) \cite{lowe2017multi} have been utilized to coordinate multiple agents in cooperative and competitive scenarios. 

\subsection{Exponential-weight algorithm for Exploration and Exploitation}
Exp3 (Exponential-weight algorithm for Exploration and Exploitation) is an adversarial bandit algorithm designed for uncertain or adversarial environments. By balancing exploration and exploitation, Exp3 minimizes regret over time \cite{exp3}. Exp3-IX refines the base Exp3 algorithm, by introducing a biased Implicit exploration toward better actions, reducing the regret’s variance, and enhancing the algorithm’s performance as uniform exploration has been shown to detrimentally impact the performance of learning algorithms, especially in environments with numerous suboptimal options \cite{exp3ix}. Exp3 and Exp3-IX provide theoretical guarantees for convergence to no-regret convergence in non-stochastic multi-armed bandit problems. 

\textit{Extending the exploration to more complex, multi-step, stochastic environments is the focus of this paper.}

\section{Algorithm}

\begin{figure}[ht]
    \centering
    \vspace{0em}
    \includegraphics[width=0.7\textwidth]{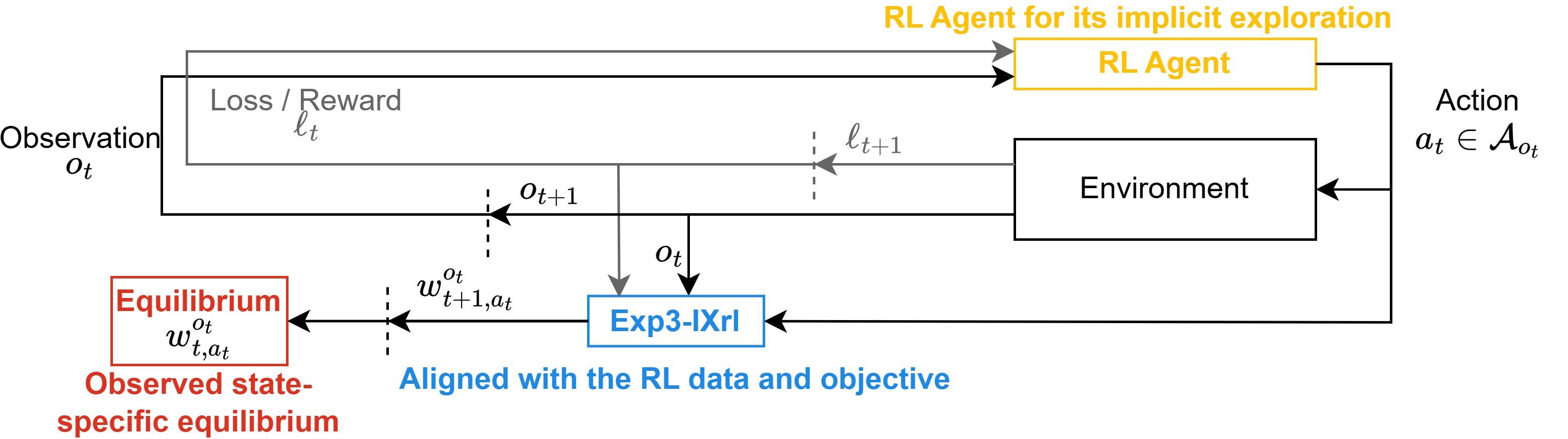}
    \vspace{-0.75em}
    \caption{\textit{Algorithm overview. Exp3-IXrl is a blend of an RL and game-theoretic approach, separating the RL agent’s action selection from equilibrium computation while preserving the integrity of the learning process.}}
    \vspace{-0.5em}
    \label{fig:algorithm_overview}
\end{figure}

We propose to extend the Exp3-IX with the reinforcement learning action selection step (see Fig~\ref{fig:algorithm_overview}). Our Exp3-IXrl incorporates several extensions to ensure compatibility across types of RL methods and asymmetric, state-specific action spaces and to enable implementation in multi-step and stochastic environments.

The action selection of an RL agent during training serves as a non-intrusive enhancement, either at each timestep $t$ or as an offline learning data, to leverage the exploration and convergence guarantees of an RL agent for concurrent CCE training. Exp3-IXrl extends the implicit exploration of the EXP3-IX algorithm, via the exploration-exploitation balances of RL algorithms, which might be more apt to handle more complex and stochastic game settings. 

At each time step during learning, EXP3-IXrl will obtain an action from either the underlying RL algorithm or the CCE approximation based on the certainty threshold. The more a state is visited, the higher the certainty in the CCE approximation.

Previous methods that attempt to approximate a CCE using EXP3 algorithms require EXP3 to interact with the environment, which differs from our proposed method. Our algorithm implements EXP3-IXrl as a third-party observer until the certainty threshold is reached. In this way, we can leverage the underlying RL algorithm's strengths to accelerate training, and once the certainty measure has been reached, we can utilize the CCE policy. A natural question is whether we lose the theoretical convergence guarantees \cite{Holtz2023} by shifting EXP3-IXrl from an active agent to a third-party observer. We introduce a normalization factor to alleviate the third-party observer effect, addressing the possible algorithm's convergence issues.
\section{Experiments and Results}

\subsection{Environments setting}

We tested Exp3-IXrl within the Cyber Operations Research Gym (CybORG) \cite{cyborg} Cage Challenge 2 environment \cite{cage_challenge_2} (CC2), a complex and adversarial cybersecurity network, where the algorithm’s objective is to minimize total network infection, represented through negative rewards. 

Moreover, we test our algorithm in a stochastic and deterministic multi-armed bandit (MAB) with ten actions. For the stochastic environment, we set the rewards to a standard normal distribution centered around zero with a standard deviation of one and add a sampled random noise to the received reward when the action is chosen. As for the deterministic environment, we set the reward based on the action number and do not add additional noise during the action selection process.

\subsection{Experimental procedure and metrics used}
To compare our algorithms to previous approaches, we train each agent for 10000 in each environment and gather the cumulative reward over the next 30 timesteps. To ensure a fair comparison of our method, we average our results over 100 runs in each environment with an exact random seed used for both the baselines and our algorithm, limiting the influence of random seeds on our results. 

Within the multi-armed bandit setting, we use classical RL algorithms as our baselines: $\epsilon$-Greedy \cite{Sutton1998}, UCB \cite{auer2002using}, and Gradient Bandit \cite{gradientBandit2005}. As for the CC2 environment, we use the CardiffUni agent, a hierarchical Proximal Policy Optimization (PPO) \cite{schulman2017proximal}, which recently had its convergence guarantees proven \cite{Holtz2023} and won the Cage Challenge 2, as our baseline.

\subsection{Results}

\begin{table}[ht]

\centering
\resizebox{\textwidth}{!}{%
\begin{tabular}{c|ccccc|}
\cline{2-6}
\multicolumn{1}{l|}{} &
  \multicolumn{5}{c|}{\cellcolor[HTML]{C0C0C0}\textbf{Algorithm}} \\ \hline
\multicolumn{1}{|c|}{\cellcolor[HTML]{C0C0C0}{\color[HTML]{333333} \textbf{Scenario}}} &
  \multicolumn{1}{c|}{\textbf{Exp3 \cite{exp3}}} &
  \multicolumn{1}{c|}{\textbf{Exp3-IX \cite{exp3ix}}} &
  \multicolumn{2}{c|}{\textbf{RL}} &
  \textbf{Exp3-IXrl} \\ \noalign{\hrule height 3pt}
\multicolumn{1}{|c|}{} &
  \multicolumn{1}{c|}{} &
  \multicolumn{1}{c|}{} &
  \multicolumn{1}{c|}{\cellcolor[HTML]{FFFFFF}{\color[HTML]{333333} \textbf{Gradient Bandit \cite{gradientBandit2005}}}} &
  \multicolumn{1}{c|}{26.49 +/- 3.58} &
  27.0 +/- 0.0 \\ \cline{4-6} 
\multicolumn{1}{|c|}{} &
  \multicolumn{1}{c|}{} &
  \multicolumn{1}{c|}{} &
  \multicolumn{1}{c|}{\cellcolor[HTML]{FFFFFF}{\color[HTML]{333333} \textbf{UCB \cite{auer2002using}}}} &
  \multicolumn{1}{c|}{22.80 +/- 0.0} &
  27.0 +/- 0.0 \\ \cline{4-6} 
\multicolumn{1}{|c|}{\multirow{-3}{*}{\textbf{\begin{tabular}[c]{@{}c@{}}Deterministic MAB [$\uparrow$] \\ Certainty: 2000\end{tabular}}}} &
  \multicolumn{1}{c|}{\multirow{-3}{*}{25.47 +/- 5.47}} &
  \multicolumn{1}{c|}{\multirow{-3}{*}{26.31 +/- 3.65}} &
  \multicolumn{1}{c|}{\cellcolor[HTML]{FFFFFF}{\color[HTML]{333333} \textbf{$\epsilon$-Greedy \cite{Sutton1998}}}} &
  \multicolumn{1}{c|}{25.77 +/- 4.75} &
  27.0 +/ -0.0 \\ \noalign{\hrule height 2pt}
\multicolumn{1}{|c|}{} &
  \multicolumn{1}{c|}{} &
  \multicolumn{1}{c|}{} &
  \multicolumn{1}{c|}{\cellcolor[HTML]{FFFFFF}{\color[HTML]{333333} \textbf{Gradient Bandit \cite{gradientBandit2005}}}} &
  \multicolumn{1}{c|}{46.37 +/- 23.31} &
  46.3 +/- 23.46 \\ \cline{4-6} 
\multicolumn{1}{|c|}{} &
  \multicolumn{1}{c|}{} &
  \multicolumn{1}{c|}{} &
  \multicolumn{1}{c|}{\cellcolor[HTML]{FFFFFF}{\color[HTML]{333333} \textbf{UCB \cite{auer2002using}}}} &
  \multicolumn{1}{c|}{48.45 +/- 21.77} &
  48.57 +/- 21.72 \\ \cline{4-6} 
\multicolumn{1}{|c|}{\multirow{-3}{*}{\textbf{\begin{tabular}[c]{@{}c@{}}Stochastic MAB [$\uparrow$] \\ Certainty: 2000\end{tabular}}}} &
  \multicolumn{1}{c|}{\multirow{-3}{*}{45.24 +/- 22.49}} &
  \multicolumn{1}{c|}{\multirow{-3}{*}{46.85 +/- 22.97}} &
  \multicolumn{1}{c|}{\cellcolor[HTML]{FFFFFF}{\color[HTML]{333333} \textbf{$\epsilon$-Greedy \cite{Sutton1998}}}} &
  \multicolumn{1}{c|}{40.24 +/- 17.69} &
  44.08 +/- 17.69 \\ \noalign{\hrule height 2pt}
\multicolumn{1}{|c|}{\textbf{\begin{tabular}[c]{@{}c@{}}CC2 [$\uparrow$] \\ Certainty: 2750  \end{tabular}}} &
  \multicolumn{1}{c|}{N / A} &
  \multicolumn{1}{c|}{N / A} &
  \multicolumn{1}{c|}{\cellcolor[HTML]{FFFFFF}{\color[HTML]{333333} \textbf{PPO (CardiffUni) \cite{schulman2017proximal}}}} &
  \multicolumn{1}{c|}{-2.94 +/- 1.41} &
  \multicolumn{1}{c|}{-3.86 +/- 1.50} \\ \hline
\end{tabular}%
}
\vspace{-1.25em}
\caption{\textit{The average cumulative reward over 30 steps averaged over 100 runs. Our algorithm significantly outperforms its classical RL and CCE counterparts.}}
\vspace{-1.25em}
\label{tab:my-table}
\end{table}

\begin{figure}[ht]
    \centering
    \includegraphics[width=0.6\textwidth]{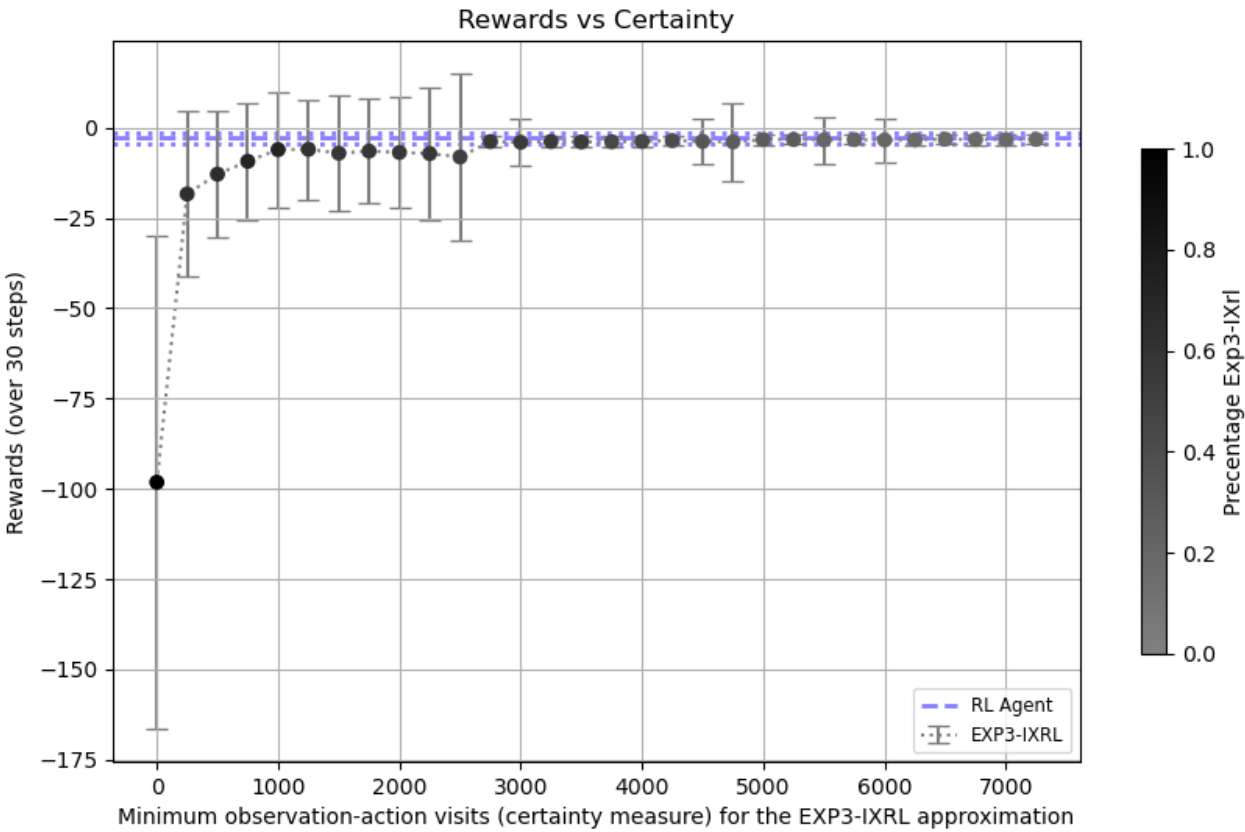}
    \vspace{-1em}
    \caption{\textit{Result of our agent in the CC2 environment with a varying certainty threshold. We achieve the performance of the PPO agent with a certainty threshold of around 2750 and with only 10000 steps, demonstrating faster convergence.}}
    \vspace{-1.25em}
    \label{fig:results}
\end{figure}

For CC2, we achieve comparable performance with a certainty threshold of 2750 in just 10000 simulation episodes - a tenth of the training episodes of the previous winning challenge submission \cite{cyborg, cage_challenge_2} (see Fig.~\ref{fig:results}). In the multi-armed bandit scenario, we illustrate our algorithm's performance against classical reinforcement learning algorithms, used as baselines and teachers for the Exp3IX-rl. Except for the Gradient Bandit algorithm in the stochastic environment, our algorithm surpasses every baseline (see Table~\ref{tab:my-table}).

\vspace{-0.75em}

\section{Conclusion and Future Work}

The proposed Exp3-IXrl algorithm combines an RL agent as an explicit exploration bias during training with traditional coarse correlated equilibrium (CCE) approximation. It maintains the RL agent’s autonomy while excelling in complex, stochastic environments, where current CCE-based implementations fail. Empirical results underscore the robustness and adaptability of Exp3-IXrl across diverse environments and policies, demonstrating enhanced learning depth.

Our findings also contribute to ongoing research about the necessity of exploration in CCE approximation, specifically the minimum certainty per action-observation pair. Future research should investigate how certainty could adjust to environmental feedback, potentially refining the RL agent’s policy and improving Exp3-IXrl’s adaptability to evolving cooperative or adversarial contexts.

\begin{footnotesize}
\bibliographystyle{unsrt}
\bibliography{references}
\end{footnotesize}

\end{document}